\documentclass[review]{elsarticle}

\biboptions{sort,compress,comma,square}

\usepackage{multicol}
\usepackage{multirow}
\usepackage{lineno}
\usepackage[english]{babel}
\usepackage[utf8]{inputenc}
\usepackage{graphicx}
\usepackage{subfigure}
\graphicspath{{./eps/}{.}}
\DeclareGraphicsExtensions{.eps}

\usepackage[cmex10]{amsmath}

\DeclareMathOperator*{\abs}{abs}
\usepackage{url}

\usepackage{setspace}

\journal{Expert Systems with Applications}
\begin{document}
\begin{frontmatter}
\title{Genetic Programming for Multibiometrics}

\author{Romain Giot\corref{corres}}
\ead{romain.giot@ensicaen.fr}
\author{Christophe Rosenberger}
\ead{christophe.rosenberger@greyc.ensicaen.fr}

\cortext[corres]{Corresponding author}
\address{GREYC Laboratory\\ ENSICAEN - University of Caen -
CNRS\\ 6 Boulevard Mar\'echal Juin 14000 Caen Cedex - France}

\begin{abstract}
Biometric systems suffer from some drawbacks: a biometric system can provide in
general good
performances except with some individuals as its performance depends highly on the
quality of the capture... One solution to solve some of these problems is to
use multibiometrics where different biometric systems are combined together
(multiple captures of the same biometric modality, multiple feature extraction
algorithms, multiple biometric modalities\ldots). In this
paper, we are interested in score level fusion functions application (i.e., we
use a multibiometric authentication scheme which accept or deny the claimant for
using an application). 
In the state of the art, the weighted sum of scores (which is a linear
classifier) and the use of an SVM (which is a non linear classifier)
 provided by different biometric
systems provid
one of the best performances. We present a new method based on the use of genetic
programming giving similar or better performances (depending on the complexity
of the database).
We derive a score fusion function by assembling some classical primitives
functions ($+$, $*$, $-$, ...).
We have validated the proposed method on three significant biometric benchmark
datasets from the state of the art.
\end{abstract}


\end{frontmatter}

\linenumbers

\section{Introduction}\label{sec:introduction}\par
\subsection{Objective}
Every day, new evolutions are brought in the biometric field of research. These
evolutions include the proposition of new algorithms with better performances, new approaches
(cancelable biometrics, soft biometrics, ...) and even new biometric modalities (like finger knuckle recognition~\cite{kumar-human}, for example).
There are many different biometric modalites, each classified among three
main families (even if we can find a more precise topology in the literature) :

\begin{itemize}
  \item \emph{biological} : recognition based on the analysis of
  biological data linked to an individual (e.g., DNA analysis~\cite{Hashiyada2004DoB}, the odor~\cite{Korotkaya2003BPA}, the
  analysis of the blood of different physiological signals, as well as heart
  beat or EEG~\cite{Riera2008UBS});

  \item \emph{behavioural} : based on the analysis of an individual behaviour 
  while he is performing a specific task (e.g., keystroke dynamics~\cite{gaines1980}, online
  handwritten signature~\cite{Fierrez2008HoB}, the way of using the mouse of the
  computer~\cite{Weiss2007MMB}, voice
  recognition~\cite{Petrovska-Delacretaz2007TSV}, gait dynamics (way of
  walking)~\cite{nandini2008comprehensive} or  way of
  driving~\cite{Benli2008DRU});

  \item \emph{morphological} based on the recognition of different particular physical
  patterns, which are, for most people, permanent and unique (e.g.,
  face recognition~\cite{turk1991face}, fingerprint
  recognition~\cite{maltoni2009hof}, hand shape
  recognition~\cite{kumar2006pru}, or blood vessel~\cite{xu5blood}, ...).
\end{itemize}

Nevertheless, there will always be some users for which a biometric modality (or
method applied to this modality) gives bad results, whereas, they are better
in average. These low performances can be implied by different facts: the quality
of the capture, the instant of acquisition and the individual itself but they
have the same implication (impostors can be accepted or user need to authenticate
themselves several times on the system before being accepted).
Multibiometrics allow to solve this problem while obtaining better
performances (i.e., better security by accepting less impostors and better
user acceptance by rejecting less genuine users) and by expecting that errors of the different modalities are not
correlated.
In this paper, we propose a generic approach for multibiometric systems.

We can find different types of biometric multimodalites~\cite{ross2006handbook}.
They use:
  \begin{enumerate}
    \item different sensors of the same biometric modality (i.e.,
    capacitive or resistive sensors for fingerprint acquisition);
    \item several different representations for the same capture
    (i.e., use of points of interest or texture for face or fingerprint
recognition);
    \item different biometric modalities (i.e., face
    and fingerprint recognition);
    \item different instances of the same modality (i.e., left
    and right eye for iris recognition);
    \item multiple captures (i.e., 25 images per second in a
    video used for face recognition);
    \item an hybrid system composed of the association of the
    previous ones.
  \end{enumerate}

  We are interested in the first four cases in this paper. 
Our objective is to
  automatically generate fusion functions which combine the scores provided by
different biometric systems in order to obtain the most efficient
multibiometrics authentication scheme.

\subsection{Background}
\subsubsection{Performance Evaluation}
In order to compare different multibiometrics systems, we need to present the
how to evaluate them.
Several works have already done on the
evaluation of biometric systems~\cite{UBESBS08,isoiec}. 
Evaluation is generally realized within three aspects:
\begin{itemize}
   \item 
 \emph{performance}: it has for objective to measure various statistical criteria on the performance of the system
(\emph{Capacity}~\cite{Bhatnagar2009}, \emph{EER}, \emph{Failure To Enroll (FTE)},
\emph{Failure To Acquire (FTA)}, computation time, \emph{ROC} curves, etc~\cite{isoiec});
    \item 
 \emph{acceptability}: it gives some information on the individuals'
\emph{perception}, \emph{opinions} and \emph{acceptance} regarding the system; %
    \item 
 \emph{security}: it quantifies how well a biometric system (algorithms
and devices) can resist to several types of logical and physical attacks such as
Denial of Service (DoS) attack.
\end{itemize}

In this paper, we are only interested in performance evaluation (because the fusion
approach is not modality dependant and perception and security depend on the
used modalities).
The main performance metrics are the following ones:
\begin{itemize}
\item \emph{FAR (False Acceptance Rate)} which represents the ratio of impostors accepted
  by the system;
\item \emph{FRR (False Rejection Rate)} which represents the ratio of genuine users
  rejected by the system;
\item \emph{EER (Error Equal Rate)} which is the error rate when the system is
  configured in order to obtain a \emph{FAR} equal to the \emph{FRR};
\item \emph{ROC (Receiver Operating Characteristic)} curve which plots the \emph{FRR} depending on the
\emph{FAR} and gives an overall overview of system performance;
\item \emph{AUC (Area Under the Curve)} which gives the area under the ROC curve.
In our case, smaller is better. It is a way to globally compare performance of
different biometric systems.
\end{itemize}

We can also present the \emph{HTER} (Half Total Error Rate) which is the mean
between the \emph{FAR} and \emph{FRR} for a given threshold (this error rate is
interesting when we cannot get the \emph{EER}).

\subsubsection{Biometric Fusion}
\begin{figure*}[!htb]
  \centering
  \subfigure[Template fusion.]{
  \includegraphics[width=.39\linewidth]{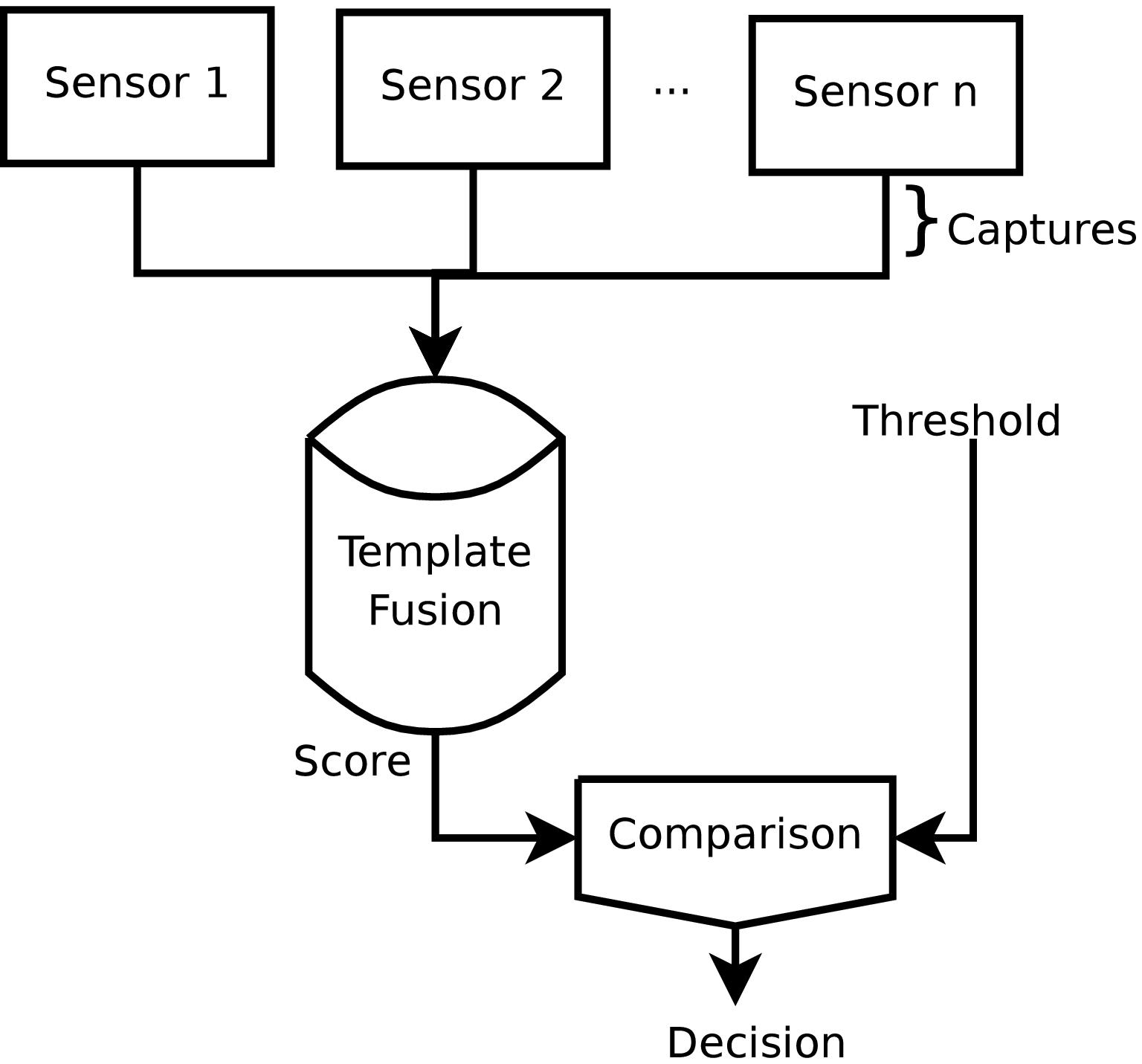}
  \label{fig:fusionmodele}
  }
  \subfigure[Classical score fusion.]{
  \includegraphics[width=.39\linewidth]{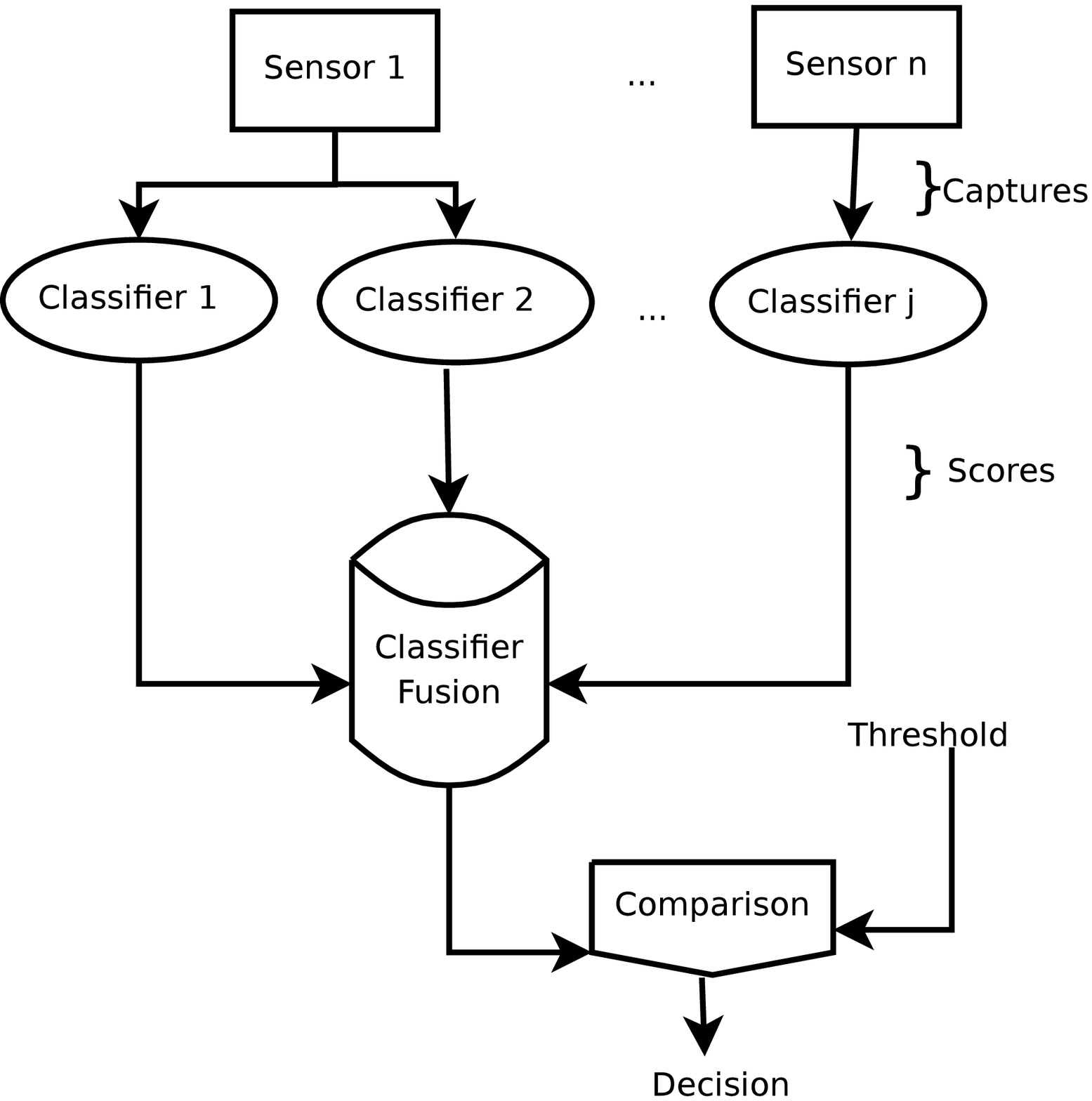}
  \label{fig:fusionscores}
  }
  \subfigure[Cascade fusion.]{
  \includegraphics[width=0.39\linewidth]{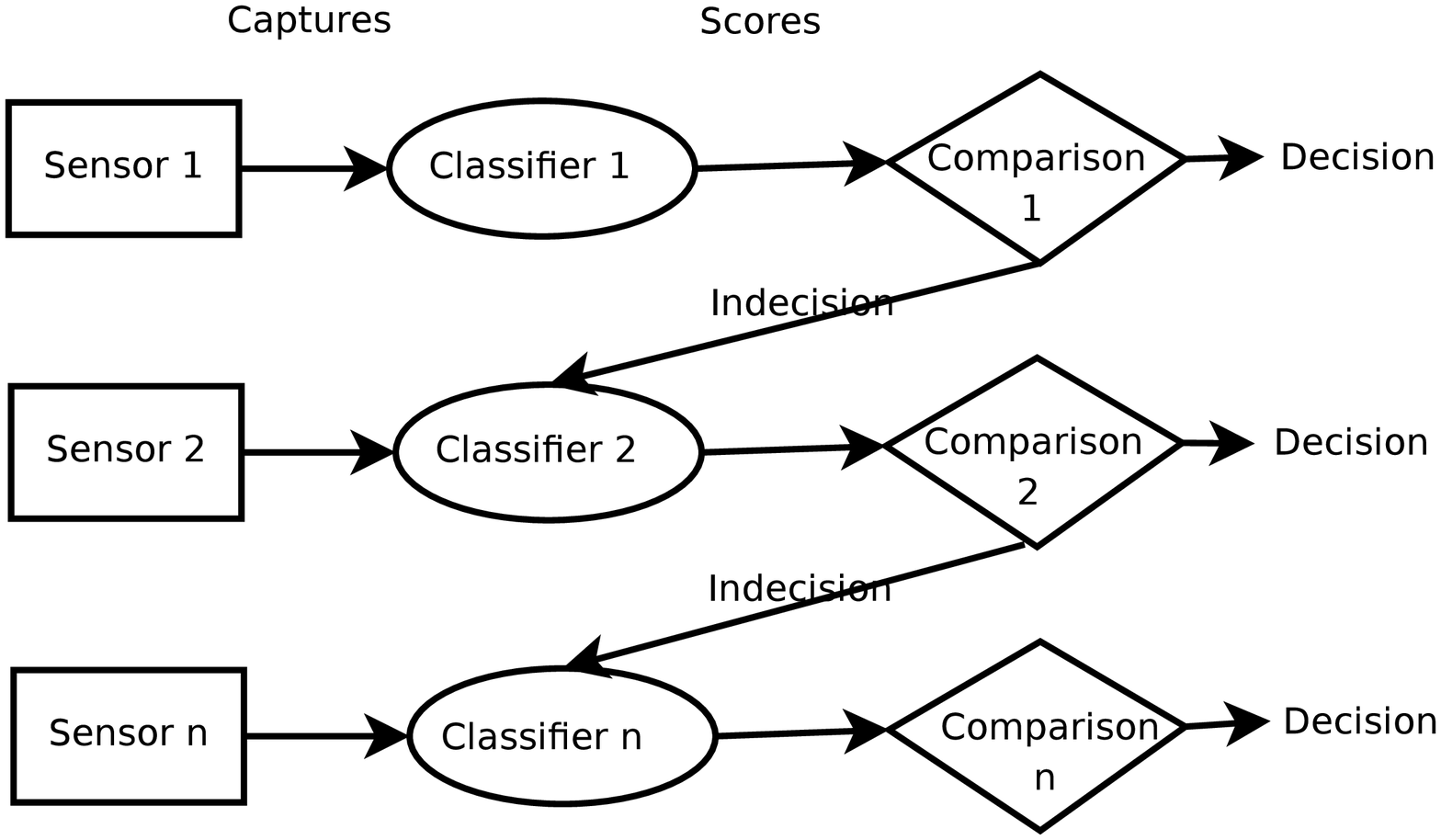}
  \label{fig:fusioncascade}
  }
  \subfigure[Hierarchical fusion.]{
  \includegraphics[width=.39\linewidth]{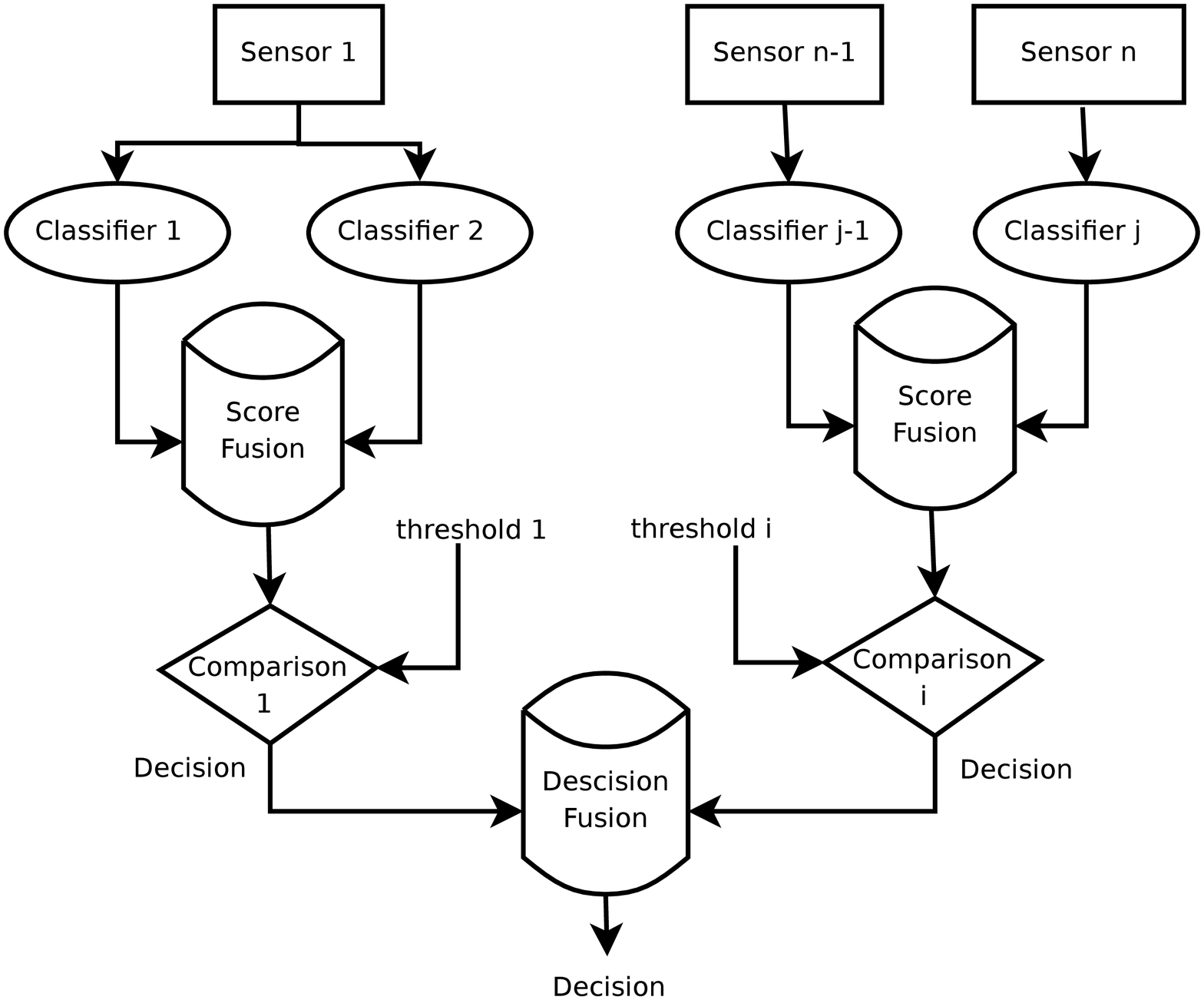}
  \label{fig:fusionhierarchique}
  }
  \caption{Illustration of different fusion mechanisms.}
  \label{fig:fusion}
\end{figure*}

There are several studies on multibiometrics. The fusion can be operated on
different points of the mechanism:
\begin{itemize}
  \item \emph{template fusion}: the templates captured by different biometric
systems are merged together, then the learning process is realized on these new
templates~\cite{Raghavendra2009PSO,rattani2009rma}.
Figure~\ref{fig:fusionmodele} presents this type of fusion. 
The fusion process is related to a feature selection in order to determine the
most significant patterns to minimize errors.

  \item \emph{decision fusion}: the decision is taken for each of the biometric
authentication system, then the final decision is done by fusing the previous
ones~\cite{ross2004multimodal}. 

	\item \emph{rank fusion}: the decision is done with the help of different ranks of
biometric identification systems. The main method is the majority vote~\cite{zuev1999voting}.

	\item \emph{score fusion}: 
the fusion is realized considering the output of the classifiers. The
Figure~\ref{fig:fusionscores} presents this type of fusion.
\end{itemize}

Buyssens \textit{et al.}~\cite{buyssens2009} showed the interest of biometric
fusion for face recognition combining the image in visible and infrared color spaces
with convolutional neural networks. 
In~\cite{montalvao2006mbf}, Mantalvao and Freire
have combined keystroke dynamics with voice recognition, it seems it is the first
time that multibiometrics has been done with keystroke dynamics and another biometric modality.
In~\cite{Hocquet2007Aba}, Hocquet \textit{et al.} demonstrated the interest of fusion in
keystroke dynamics in order to improve the recognition rates: three different
keystroke dynamics functions are used on the same capture. The sum operator
(consisting in summing the different scores)
seems to be the most powerful approach in the literature.

These fusion architectures are quite simple but powerful. Results can yet be improved
(in term of error rate or computation time) by using different architectures.
A cascade fusion~\cite{allano2009} is another interesting approach. A first test
is done, if the user is correctly verified as the attended client or if it is
detected as an impostor, the algorithm stops. Otherwise, another biometric
authentication (with another capture from another modality) is proceeded until
obtaining a decision of acceptance or rejection, or reaching the end of the
cascade. So, instead of using one decision threshold, each test (except the
last one) needs two thresholds: one for rejection and one for acceptance. All scores
between these thresholds are considered in an indecision zone. This mechanism is
presented in Figure~\ref{fig:fusioncascade}. 
Another advantage of this method is to decrease the verification time by not using
all the modalities, they are used only if necessary. This method has been
successfully applied on a multibiometric system using face
and fingerprint recognition in a mobile environment (where acquisition
and computation times are important)~\cite{allano2009}.

Another kind of architecture has been proposed: it is a hierarchical fusion
scheme~\cite{shen2009aml} (called multiple layers by their authors).
Shen \textit{et al.} have presented this method with two different keystroke
dynamics methods. The fusion is done at different steps, and involves different
mathematical operations on scores
(sum, weighted sum, product, min, max) and logical operations
decision (comparison to a threshold, or, and) on differents templates extracted
from the same capture. An extended version to any multibiometric system is
presented in Figure~\ref{fig:fusionhierarchique}. We think our work can be seen
as a generalization of this paper.\\

It is also possible to model the distribution of the genuine and impostor
matching scores, we talk about \emph{Density-based score
fusion}. In~\cite{nandakumar2008likelihood}, scores are modelled with a Gaussian
Mixture Model and have been tested on three multibiometric databases involving
face, fingerprint, iris and speech modalities.\\

Concerning non linear algorithms, Support Vector Machine (SVM) can also be used in a fusion process. Each score to combine is
arranged in a vector and a training set is used to learn the SVM model.
In~\cite{Czyz}, the SVM fusion to improve face recognition gives slightly better performances than weighted sum.
Voice and online signature have been fused with SVM in~\cite{garcia2005multimodal}.
In this experiment, arithmetic mean gives best results with noise free data, while
SVM gives equivalent results with noisy data.

\subsection{Discussion}
In this paper, we are interested in biometric modality independent
\emph{transformation-based score fusion}~\cite{nandakumar2008likelihood} where
the matching scores are first normalized and second combined.
We have previously seen that in this case, arbitrary functions are often used.
Our work is based on these various fusion architectures based on score
fusion in order to produce a score fusion function automatically generated with genetic
programming~\cite{koza1992genetic}. \\

By the way, the definition of a fusion architecture is still an open
issue in the multibiometrics research field~\cite{ross2009mso}, because the range of possible fusion configurations
is very large. We think that using automatically generated fusion functions can bring a
new solution to solve this kind of problems.


\section{Material and Methods}

In this section, we present all the required information in order to allow other
researchers to reproduce our experiment.

\subsection{Biometric databases}
As it is well known that results can be highly related to the database,
for this study, we have used three different multibiometric databases: the first
one is the BSSR1~\cite{NISTBSSR} distributed by the
NIST~\cite{bssr1url}
(referenced as BSSR1 in the paper), the second one is a
database we have created for this purpose (referenced as PRIVATE in the
paper) and the third one is a subset of scores computed with the
BANCA~\cite{bailly2003banca} database
(referenced as BANCA in the text. In fact, BANCA database is composed of
templates. We have used the scores available in~\cite{bancaurl}).
As all these databases are multi-modal, the scores are presented with tuples:
the $i$\emph{{th}} tuple of scores is represented as $s_i = (s_i^1, s_i^2, ..., s_i^n)$ for
a database having $n$ modalities (in our case, $n \in \{4,5\}$).

The three databases are presented in detail in the following subsections while
Table~\ref{tab:databases} presents a summary of their description.

\subsubsection{BSSR1 database}
The BSSR1~\cite{NISTBSSR} database consists of an ensemble of scores sets from different
biometric systems. In this study, we are interested in the subset containing the
scores of two facial recognition systems and the two scores of a fingerprint
recognition system applied to two different fingers for $512$ users.
We have $512$ tuples of intra-scores (comparison of the capture of an individual
with its model) and $512*511=261,632$ tuples of inter-scores (comparison of the
capture of an individual with the model of another individual). Each tuple is
composed of 4 scores:  $s = (s^1_{bssr1}, s^2_{bssr1}, s^3_{bssr1}, s^4_{bssr1})$, 
they respectively represent  the score of
the algorithm A of face recognition, the score of algorithm B of face
recognition (the same face image is used for the two algorithms), 
the score of the fingerprint recognition with left index, the score
of fingerprint recognition with right index.
This database has been used several times in the 
literature~\cite{nandakumar2008likelihood,sedgwick2005preliminary}.

\subsubsection{PRIVATE database}
The second database is a chimeric one we have created by
combining two public biometric template databases: the AR~\cite{martinez1998ar}
for the facial recognition and the GREYC keystroke~\cite{giot2009benchmark} for
keystroke dynamics.\\

The AR database is composed of frontal facial images of $126$ individuals under
different facial expression, illumination conditions or occlusions. This is a
quite difficult database in reason of these specificities. These images
have been taken during two different sessions with $13$ captures per session.
The GREYC keystroke contains the captures on several session during a two months
period involving $133$ individuals. Users were asked to type the password "greyc laboratory" $6$
times on a laptop and $6$ times on an USB keyboard by interleaving the typings.

We have selected the first $100$ individual of the AR database and we have
associated each of these individuals to another one in a subset of the GREYC
keystroke database having $5$ sessions of captures. We then used the $10$ first
captures to create the model of each user and the $16$ remaining ones to compute the
intra and inter scores.

These scores have been computed by using two different methods for the face
recognition (the scores $s^1_{private}$ and $s^2_{private}$ and three different ones for the
keystroke dynamics ($s^3_{private}$, $s^4_{private}$ and $s^5_{private}$ scores). The face
recognition algorithms are based on eigenfaces~\cite{turk1991face} and SIFT
keypoints~\cite{lowe2004distinctive} comparisons between images from the model and the
capture~\cite{CI-ROSENBERGER-2008}. Keystroke dynamics scores have been computed
by using different methods~\cite{CI-GIOT-2009-2} based on SVM, statistical
information and rhythm measures.

\subsubsection{BANCA database}
The lastest used benchmark is a subset of scores produced by the help of the BANCA
database~\cite{bancaurl}. The selected scores  correspond to the following one labelled:
IDIAP\_voice\_gmm\_auto\_scale\_25\_100\_pca.scores for $s^1_{banca}$,
SURREY\_face\_nc\_man\_scale\_100.scores for $s^2_{banca}$,
SURREY\_face\_svm\_man\_scale\_0.13.scores for $s^3_{banca}$ and
\linebreak UC3M\_voice\_gmm\_auto\_scale\_10\_100.scores for $s^4_{banca}$.

We have empirically chosen this subset.
G1 set is used as the learning set, while G2 set is used as the validation set.
Users from G1 are different than users from G2.

\subsubsection{Discussion}

The main differences between these three benchmarks are:
\begin{itemize}
  \item the biometric modalities used in BSSR1 and BANCA have better performances than the
  ones in PRIVATE;
  \item the quantity of intra-scores is more important in PRIVATE (only one
  tuple of intra-score per user in BSSR1 instead of several in PRIVATE);
  \item BSSR1 and BANCA are databases of scores (by the way, we do not know
 the biometric systems having generated them) whereas PRIVATE is a database of templates (we had to
  compute the scores);
  \item BSSR1 and BANCA are more adapted to physical access control applications (i.e., a building is protected
  by a multi-modal biometric system), while PRIVATE is more adapted to logical
  access control (i.e., the authentication to a Web service is protected by a multi-modal biometric system).
\end{itemize}

\begin{table}[!tb]
  \caption{Summary of the different databases used to validate the proposed method}
  \small
  \label{tab:databases}
  \centering
  \begin{tabular}{|l|r|r|r|}\hline
    \textbf{Nb of} & \textbf{BSSR1} & \textbf{PRIVATE} & \textbf{BANCA}\\\hline\hline
    users & 512 & 100 & 208 \\ \hline
    intra tuple & 512 & 1600 & 467 \\ \hline
    inter tuple & 261632 & 158400 & 624 \\ \hline
    items/tuples  & 4 & 5 & 4 \\ \hline
\end{tabular}
\end{table}

In the following subsections, we describe the proposed methodology to
automatically generate a score fusion function with genetic programming. We
adopt the classical score fusion context described in Figure~1(b). Before
using the scores provided by different biometric systems, we need to normalize
them.

\subsection{Score Normalization}
It is necessary to normalize the various scores before operating the fusion
process: indeed, these scores come from different classifiers and their values
do not necessarily evolve within the same interval.
We have chosen to use the \emph{tanh}~\cite{hampel1986robust} operator to
normalize the scores of each modality. Equation (\ref{eq:tanh}) presents the
normalization method, where $\mu_{gen}^m$ and $\sigma_{gen}^m$
respectively represents the average and standard deviation of the genuine scores
of the modality $m$. The genuine scores are obtained by comparing the model and
the capture of the same user: they are also called the \emph{intra scores}. In
opposition, the \emph{inter scores} are obtained by comparing the model of a user with
the capture of other users. $score'$ and $score$ respectively represents the scores
after and before normalisation.

\begin{equation}
  \label{eq:tanh}
  score'= \frac{1}{2}\left\{tanh\left(\frac{1}{100}(\frac{score-\mu_{gen}^m}{\sigma_{gen}^m}\right) +1 \right\}
\end{equation}

We have selected this normalization procedure from the state of the art because it is known to be
stable~\cite{Jain20052270} and does not use impostors patterns which can be hard
or impossible to obtain in a real application. The aim of this paper is not to analyse the performance
of biometric systems depending on the normalization procedure, but to present a
new multibiometrics fusion procedure.
The scores of each modality have been normalized using
this procedure.

\subsection{Fusion Procedure}

In this study, we have chosen to use genetic programming~\cite{koza1992genetic}
 in order to generate score fusion functions. Genetic programming belongs to the family of
evolutionary algorithms and its scheme is quite similar to the one of genetic
algorithms~\cite{mitchell1998introduction}: a population of computer programs
(possibly represented by a tree) evolves during several generations;
different genetic operators are used to create the new population.
Programs are evaluated by using a fitness function which produces a value that
is used for their comparisons and gives a probability of selection during the
tournaments.
In a system where the computer programs are represented by trees, their
leaves mainly represent the entries of the problem, the root gives the solution
to the problem and the other nodes are the various functions taking into
arguments the values of their children nodes.

The leaves are called terminals and can be of several kinds: (a) pseudo-variables
containing the real entries of the problem (in our case, the list of scores of
each modality), (b) some constants possibly randomly generated, (c) functions
without any arguments having any side effect, or (d) some ordinary variables.

\begin{figure}[!tb]
\centering
\subfigure[Program source 1]{\includegraphics[width=0.4\linewidth]{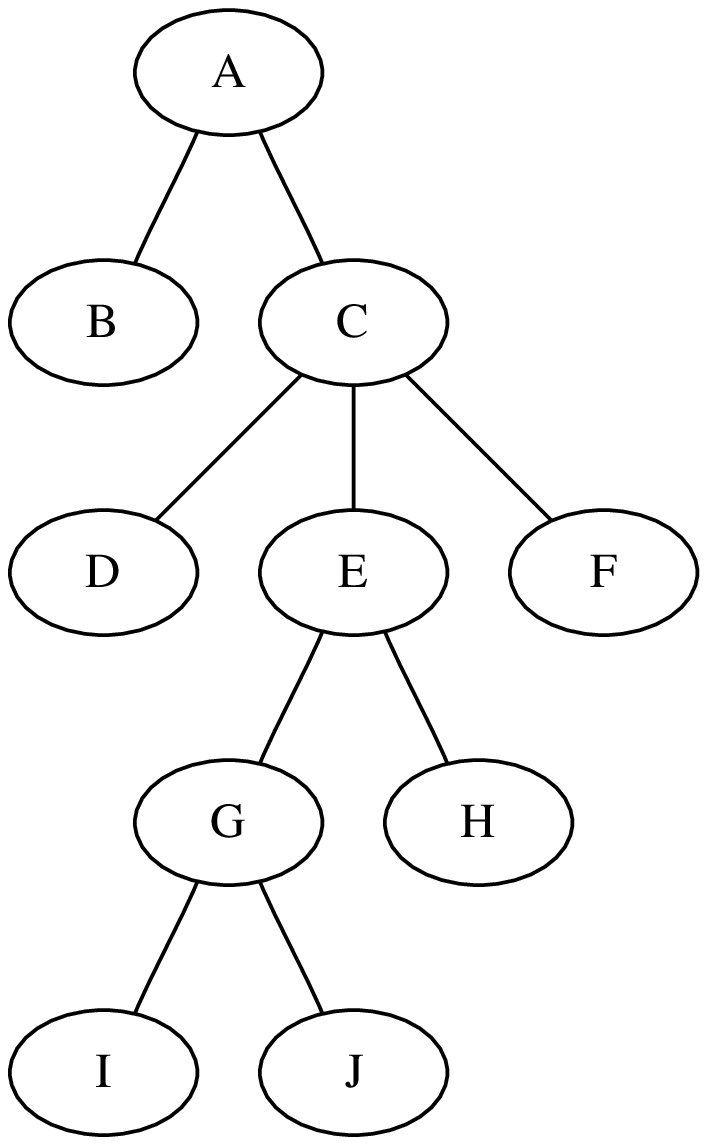}}
\subfigure[Program source 2]{\includegraphics[width=0.4\linewidth]{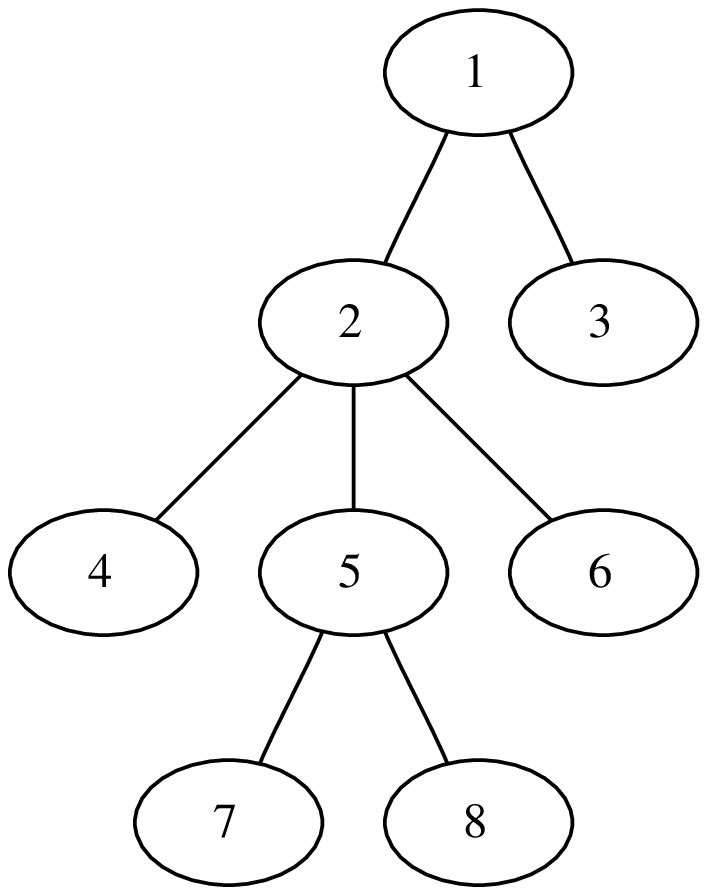}}
\subfigure[Program result 1]{\includegraphics[width=0.4\linewidth]{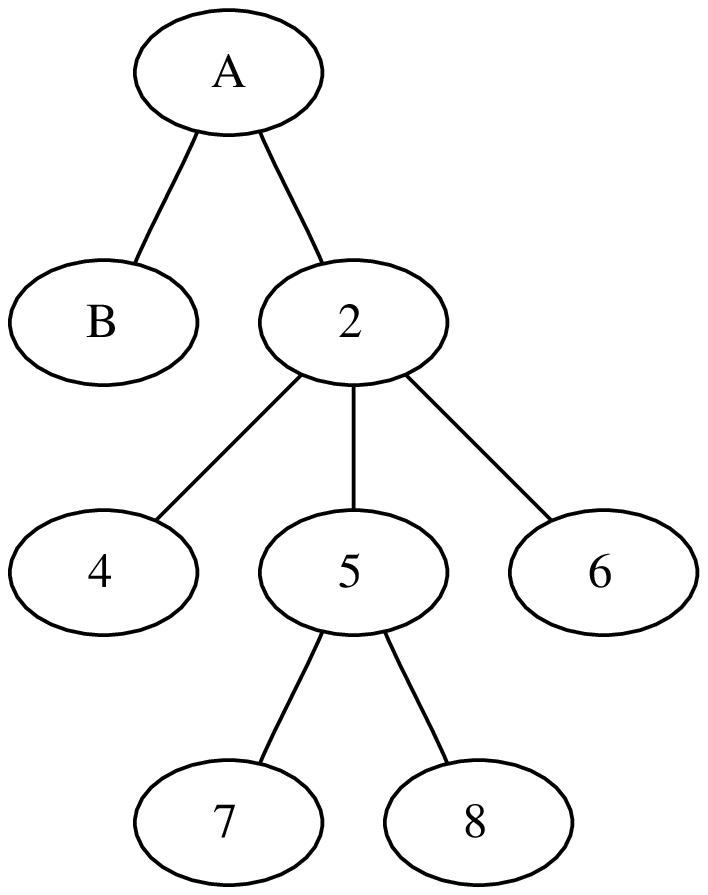}}
\subfigure[Program result 2]{\includegraphics[width=0.4\linewidth]{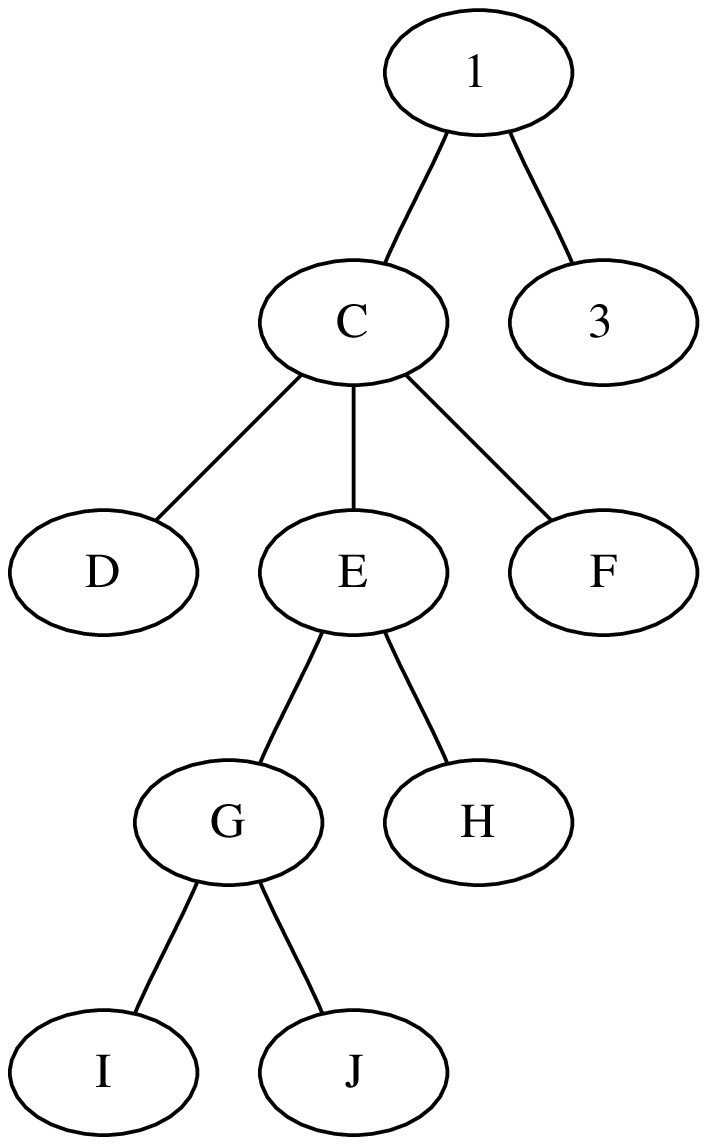}}
\caption{Crossover in genetic programming: node C from tree 1 is exchanged with
node 2 from tree 2. Program result 1 is the new individual to add to the new
generation.}
\label{fig:crossover}
\end{figure}

\begin{figure}[!tb]
\centering
\subfigure[Program source]{\includegraphics[width=0.3\linewidth]{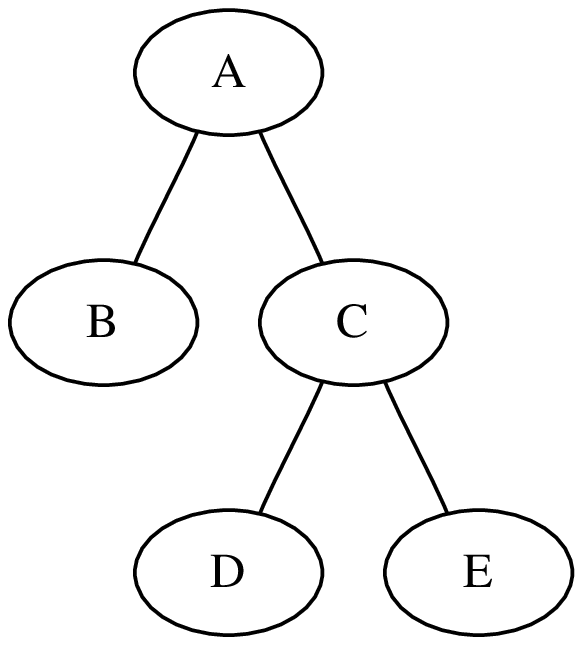}}
\subfigure[Program result]{\includegraphics[width=0.6\linewidth]{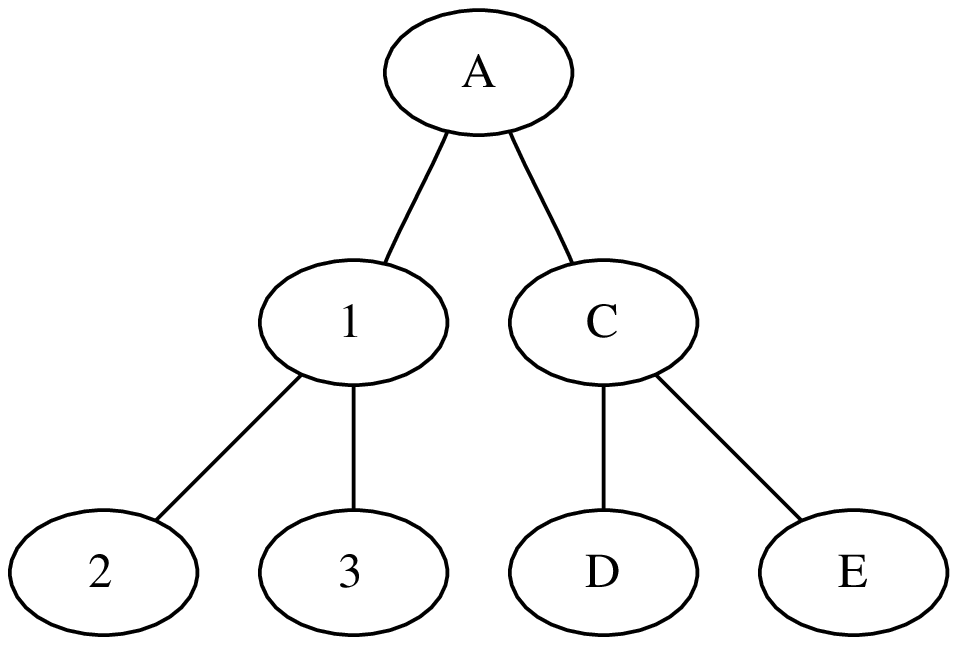}}
\caption{Mutation in genetic programming: node B is replaced by another sub-tree.}
\label{fig:mutation}
\end{figure}

The different genetic operators usually used during the evolution are (a) the
crossover, where
randomly choose sub-trees have two different trees are exchanged, (b) the mutation,
where a sub-tree is destroyed and replaced by another one randomly generated, or
(c) the copy, where the tree is conserved in the next generation.
The different steps of a genetic programming engine are presented as following:
\begin{enumerate}
  \item An initial population is randomly generated. This population is composed
of computer programs using the available functions and terminals. The trees are
built using a recursive procedure.

  \item The following steps are repeated until the termination criterion is satisfied
(the fitness function has reached the right value, or we reached the maximum
number of generations).
  \begin{enumerate}
    \item Computation of the fitness measure of each program (the programming is
evaluated according to its input data).
    \item Selection of programs with a probability based on their fitness to apply
them the genetic operations.
    \item Creation of the new generation of programs by applying the following
genetic operations (depending on their probabilities) to the previously selected programs:
     \begin{itemize}
        \item Reproduction: the individual is copied to the new population.
	\item Crossover: A new offspring program is created by recombining
	randomly chosen parts from two select programs. An example is provided in Figure~\ref{fig:crossover}.
	\item Mutation: A new offspring program is created by mutating one node
	of the selected program at a randomly chosen place. An example is provided in Figure~\ref{fig:mutation}.
     \end{itemize}
  \end{enumerate}
  \item the single best program of the whole population is designated as the
  winner. This can be the solution or an approximate solution to the problem.
\end{enumerate}~

Different applications to genetic programming are presented
in~\cite{poli2008field} as well as their bibliographic references. The fields of
these applications can be listed in curve fitting, data modelling, symbolic
regression, image and signal processing, economics, industrial process control,
medicine, biology, bioinformatics, compression... but, it seems, so far of our
knowledge, that it has not been yet applied to 
multibiometrics. We only found one reference on genetic programming in the
biometrics field. In this paper~\cite{day2007rts}, authors have used genetic programming to
learn speaker recognition programs. They have used an island model where
different islands operate their genetic programming evolution, and, after each
generation some individuals are able to leave to another island. The obtained
performance was similar to the state of the art in speaker recognition in normal
conditions, but, the generated systems performed better in degraded conditions.

More information about the configuration of the genetic programming system is
presented in the next section.

\subsection{Parameters of the Genetic Programming}
 We want to use a score fusion function that returns a score related to the
performance of a
multibiometric system. This score has to be compared with a threshold in order
to make the decision of acceptance or rejection of the user. 
In this case, none logical operation is required in the generated programs and
different information can be extracted from the result of the fusion function
(we can compute the ROC curve, the EER, ...).

\subsubsection{Fitness Function}
The EER (Error Equal Rate) is usually used to compare the performance of
different biometric systems together. A low EER means that FAR and FRR are both
low and the system has a good performance if its threshold is configured
accordingly to obtain this value. For this reason, we have chosen to use this running point to evaluate the
performance of the generated score fusion functions.

To compute the EER, we consider the highest and lowest values in the final scores
generated by the genetic programming. Then, we set a threshold at the lowest score and
linearly increment it until obtaining the highest score value in 1000 steps.
For each of these steps, we compute the FAR (comparison between the threshold
and the inter scores) and FRR (comparison between the threshold and the
intra scores). 
The ROC curve can be obtained by plotting all these couples of
(FAR, FRR), while the EER is the mean of FAR and FRR for the couple having the
lowest absolute difference. So, the fitness function is $fitness=(FAR_i+FRR_i)/2$, 
where $i$ is the threshold for which $\abs(FAR_i-FRR_i)$ is minimal.

\subsubsection{Genetic Programming Parameters}
In this section, we present the various parameters used in the genetic
programming algorithm.
Table~\ref{tab:gpparams} presents the various parameters of the evolutionary
algorithm.
\begin{table*}[!tb]
	\singlespace
  \caption{Summary of the configuration of the genetic programming iterations.
  Numbers used in function set can be scores or constants.}
  \label{tab:gpparams}
  \centering
 \begin{tabular}{|p{3cm}|p{12cm}|}\hline
   
   \textbf{Configuration} & 
   \textbf{Values}
   \\\hline\hline

   \textbf{Objective} &
	Generates a function producing a multibiometrics score.
   \\\hline

   \textbf{Functions set} & 
	\setlength{\columnsep}{1em}
	\begin{multicols}{2}
	\begin{itemize}\setlength{\itemsep}{-2mm}
		\item $+$: addition of two numbers, 
		\item $-$: subtraction of two numbers,
		\item $*$: multiplication of two numbers, 
		\item $/$: division of two numbers, 
		\item $min$: returns the minimum of two numbers,
		\item $max$: returns the maximum of two numbers,
		\item $avg$: returns the mean of two numbers
	\end{itemize}
	\end{multicols}
  \\\hline

  \textbf{Fitness function} & 
        Computes the EER of the multibiometric system
  \\\hline\hline

  \textbf{Terminal set} & 
	\setlength{\columnsep}{0.5em}
	\begin{multicols}{3}

	\textbf{BSSR1}
        \begin{itemize}\setlength{\itemsep}{-2mm}
          \item $a$: scores from $s^1_{bssr1}$,
          \item $b$: scores from $s^2_{bssr1}$,
          \item $c$: scores from $s^3_{bssr1}$, 
          \item $d$: scores from $s^4_{bssr1}$,
          \item 50 constants linearly distributed between 0 and 1.
        \end{itemize}\pagebreak
        
	\textbf{PRIVATE}
        \begin{itemize}\setlength{\itemsep}{-2mm}
          \item $a$, $b$, $c$: keystroke dynamics scores ($s^3_{private}$, $s^4_{private}$, $s^5_{private}$),
          \item $d$, $e$: face recognition scores ($s^1_{private}$, $s^2_{private}$),
          \item 50 constants linearly distributed between 0 and 1.
        \end{itemize}

	\textbf{BANCA}
        \begin{itemize}\setlength{\itemsep}{-2mm}
	  \item $a$: scores from $s^1_{banca}$,
          \item $b$: scores from $s^2_{banca}$,
          \item $c$: scores from $s^3_{banca}$, 
          \item $d$: scores from $s^4_{banca}$,
          \item 50 constants linearly distributed between 0 and 1.
        \end{itemize}

\end{multicols}
    \\\hline


  \textbf{Initial population}& 
  500 random trees with a depth between 2 and 8 built with the ramped half and
  half method.
  \\\hline

  \textbf{Evolution parameters} & 
	\setlength{\columnsep}{1em}

	\begin{multicols}{2}    \begin{itemize}\setlength{\itemsep}{-2mm}
      \item Number of individuals: 500, 
      \item Maximal number of generations: 50, 
      \item Depth limited to: 8, 
      \item Probability of crossover: 45\%,
      \item Probability of mutation: 50\%
      \item Probability of reproduction: 5\% (with elitism),
      \item Selection: tournament of size 10 with a selection probability of 80\%.
    \end{itemize}
	\end{multicols}
    \\\hline
  
    \textbf{Termination criterion} & 
    Best individual has a fitness inferior at 0.001 (by the way, this value
    would never be met \dots) or maximal number of generations reached.
    \\\hline

    \textbf{Learning set} &
    First half of the intra-scores tuples and first half of the inter-scores tuples.
    \\\hline
    \textbf{Validating set} & 
    Second half of the intra-scores tuples and second half of the inter-scores tuples.
    \\\hline
  \end{tabular}
\end{table*}

To achieve this experiment, we used the PySTEP~\cite{pySTEP} library.
The generated programs contain basic functions ($+$, $-$, $*$, $/$, $min$, $max$,
$avg$). The terminals are the scores of the biometric systems and random
constants between 0 and 1. 

The whole fitness cases are completed with a single
tree evaluation, thanks to the numpy~\cite{oliphant2006guide} library.
Each fitness case is a tuple of scores (where each score comes from a different
biometric modality) and its result value is the score returned by the generated multimodal system. The global
fitness value of a tree is the EER value computed with the previously generated
scores (computation of the ROC curve, then reading of the EER value from it).

PySTEP is a strongly typed genetic programming engine, but, in our case,
we do not use any particular constraints: the root node can only have a function
as child (no terminal in order to avoid an unimodal system, and any function of
the set), while the other
function nodes can have any of the functions as children as well as any of
the terminals.

The maximal depth of the generated trees is set to 8.
In order to avoid to stay in a local minimal solution, the mutation probability is set to
50\%.
500 individuals evolve during 50 generations. 
We have set this few
quantities, because during our investigations, using a population of 5000
individuals on 100 generations did not give so much better results (gain not
interesting in comparison to the computation time). Each database has been splitted in two
sets of equal size: the first half is the learning set and the second half is
the validation set.

The mutation rate is set to 50\%, the cross-over rate to 45\% and the
reproduction rate to 5\%. For mutation and cross-over the individuals are
selected with a tournament of size 10 with a probability of 80\% to select the
best individual. The same individual can be selected several times. For the
reproduction, the individuals are selected with an elitism scheme: the 5\% best
individuals are copied from generation $n-1$ to generation $n$. During a
crossover, only the first offspring (of the two generated ones) is kept.

\section{Results}
\label{sec:res}

In this section, we present the results of the generated fusion programs
on the three benchmark data sets. 

The results are compared to other functions
from the state of the art: (a) the $min$ rule which returns the minimum score
value, (b) the $mul$ rule which returns the product of all the scores, (c) the
$sum$ rule which returns the sum of the scores, (c) the $weight$ rule
which returns a weighted sum, and (d) an SVM implementation.
The weighs of the weighted sum have been configured by using genetic algorithm
on the training sets~\cite{CI-GIOT-2010,CI-GIOT-2010-2} (in order to give the best results as possible).
The fitness function is the value of the EER and the genetic algorithm engine
must lower this value.
Table~\ref{tab:ga} presents the configuration of the genetic algorithm.

\begin{table}[!tbh]
\caption{Configuration of the genetic algorithm to set the weights of the
weighted sum}
\label{tab:ga}
\begin{tabular}{|l|p{2.7cm}|}\hline
\textbf{Parameter} & \textbf{Value}\\\hline\hline
Population & 5000  \\\hline
Generations & 500   \\\hline
Chromosome signification & weights of the fusion functions \\\hline
Chromosome values interval & $[-10;10]$ \\\hline
Fitness    & EER on the generated function\\\hline
Selection  &  normalized gemetric selection (probability of 0.9)\\\hline
Elitism & True \\\hline
\end{tabular}
\end{table}

For the SVM, we have computed the best parameters (i.e., search the $C$ and
$\gamma$ parameter giving the lowest error rate) using the learning database on a 5-fold cross validation
scheme. We have used the $easy.py$ script provided with libSVM~\cite{chang2001libsvm} for this
purpose. We have then tested the performance on the validation set. We only
obtain on functional point (and not a curve) when using an SVM. That's why we
have used the HTER instead of the EER.

Table~\ref{tab:initperf} presents the performances, for the three databases,
of each biometric systems, fusion mechanisms from the sate of the art, and our
contribution.

Concerning the state of the art performances, can see that the simple fusion functions $sum$ and $mul$ tend to give better
performances compared to the best biometric method of each database, but they are
outperform by the $weight$ rule. The $min$
operator gives quite bad results (it does not improve the best biometric
system). The $SVM$ method gives good results but is outperform by the $weight$
method.

\begin{table}[!tb]
  \caption{Performance (HTER in \%) of the initial methods ($s^1_*$, $s^2_*$,
$s^3_*$, $s^4_*$, $s^5_*$), the state of the art fusion
  functions ($sum$, $min$, $mul$, $weight$) and our proposal on the three databases. Bold values
represent better performance than the initial biometric systems, and * represents fusion results better than state of the art.}
  \small
  \label{tab:initperf}\centering

	\subfigure[BSSR1]{
  \begin{tabular}{|ll|c|}\hline
    \multicolumn{2}{|l|}{\textbf{Method}} & \textbf{HTER} \\\hline
   
    \multicolumn{3}{|c|}{\textbf{BSSR1}}\\\hline
\multirow{4}{*}{Biometric systems} 
&$s^1_{bssr1}$ &04.30\% \\
&$s^2_{bssr1}$ &06.19\% \\
&$s^3_{bssr1}$ &08.41\% \\
&$s^4_{bssr1}$ &04.54\% \\\hline
\multirow{5}{*}{Fusion functions} 
&$sum$& \textbf{00.70\%}\\
&$min$& 05.04\% \\
&$mul$& \textbf{00.70\%}\\
&$weight$ & \textbf{00.38\%} \\ 
&$SVM$ & \textbf{0.77\%}  (FAR=1.16\%, FRR=0.39\%)\\\hline
Proposal & gpI& \textbf{0.40\%} \\ 
\hline

  \end{tabular}}

  \subfigure[PRIVATE]{
  \begin{tabular}{|ll|c|}\hline
    \multicolumn{2}{|l|}{\textbf{Method}} & \textbf{HTER} \\\hline
   
\multicolumn{3}{|c|}{\textbf{PRIVATE}}\\\hline
\multirow{4}{*}{Biometric systems} 
&$s^1_{private}$ &8.92\% \\
&$s^2_{private}$ &11.53\% \\
&$s^3_{private}$ &15.69\% \\
&$s^4_{private}$ &06.21\% \\
&$s^5_{private}$ &31.43\% \\\hline
\multirow{5}{*}{Fusion functions} 
&$sum$&\textbf{02.70\%}\\
&$min$&13.72\% \\
&$mul$&\textbf{02.67\%}\\
&$weight$ & \textbf{02.26\%} \\ 
&$SVM$ & \textbf{05.47\%} (FAR=10.87, FRR= 0.07\%) \\\hline
Proposal &    gpA&  \textbf{01.57\%}*\\ 

\hline
  \end{tabular}}

  \subfigure[BANCA]{
  \begin{tabular}{|ll|c|}\hline
    \multicolumn{2}{|l|}{\textbf{Method}} & \textbf{HTER} \\\hline

\multicolumn{3}{|c|}{\textbf{BANCA}}\\\hline
\multirow{4}{*}{Biometric systems} 
&$s^1_{banca}$ &04.38\% \\
&$s^2_{banca}$ &11.54\% \\
&$s^3_{banca}$ &08.97\% \\
&$s^4_{banca}$ &07.32\% \\\hline
\multirow{5}{*}{Fusion functions} 
&$sum$&\textbf{01.28\%}\\
&$min$&04.38\% \\
&$mul$&\textbf{01.28\%}\\
&$weight$& \textbf{00.91\%} \\ 
&$SVM$ & \textbf{01.01\%} (FAR= 1.71 \%, FRR=0.32\%) \\\hline
Proposal &   gp$\Phi$& \textbf{00.75\%}*\\ 
\hline
 
\end{tabular}}

\end{table}

Table~\ref{tab:res2} presents the gain of performance against the $weight$ operator (which
gives the best results in Table~\ref{tab:initperf}) in term of EER and AUC.

\begin{table}
\centering
\caption{Performance gain betwain our proposal and the weighted sum (which gives
the best results in the methods of the state of the art).}
\label{tab:res2}
\begin{tabular}{|l|c|c|} \hline
\textbf{Database} & \textbf{EER} & \textbf{AUC} \\ \hline

BSSR1  & -5.26\%& \textbf{0.05\%}\\\hline

PRIVATE & \textbf{34.85\%} & \textbf{23.85\%}\\\hline
BANCA& \textbf{17.58\%} & \textbf{76.74\%} \\\hline
\end{tabular}
\end{table}

\begin{figure*}[!tbh]
  \centering
  \subfigure[Validation with BSSR1]{\includegraphics[width=.7\linewidth]{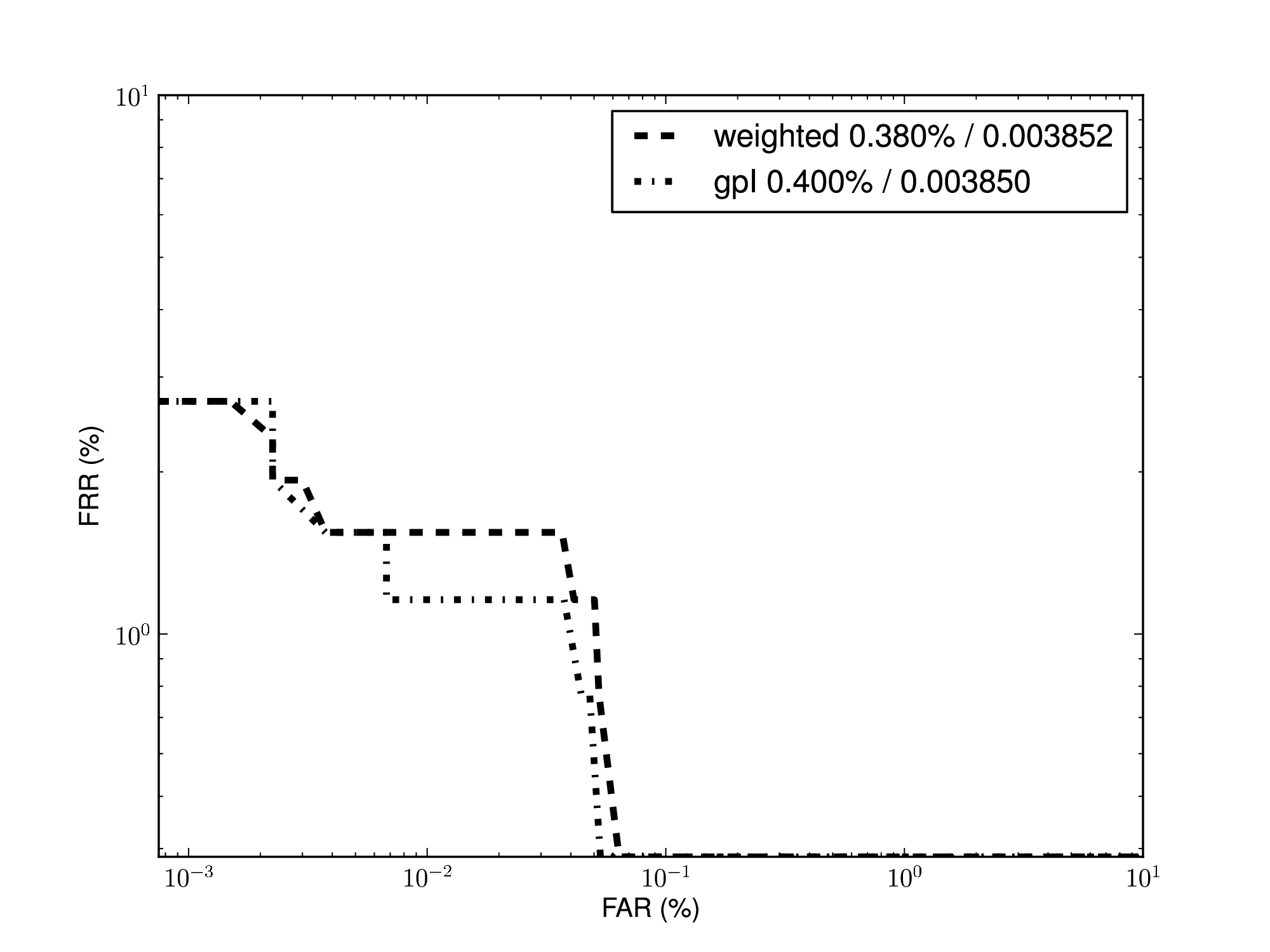}}

  \subfigure[Validation with PRIVATE]{\includegraphics[width=.7\linewidth]{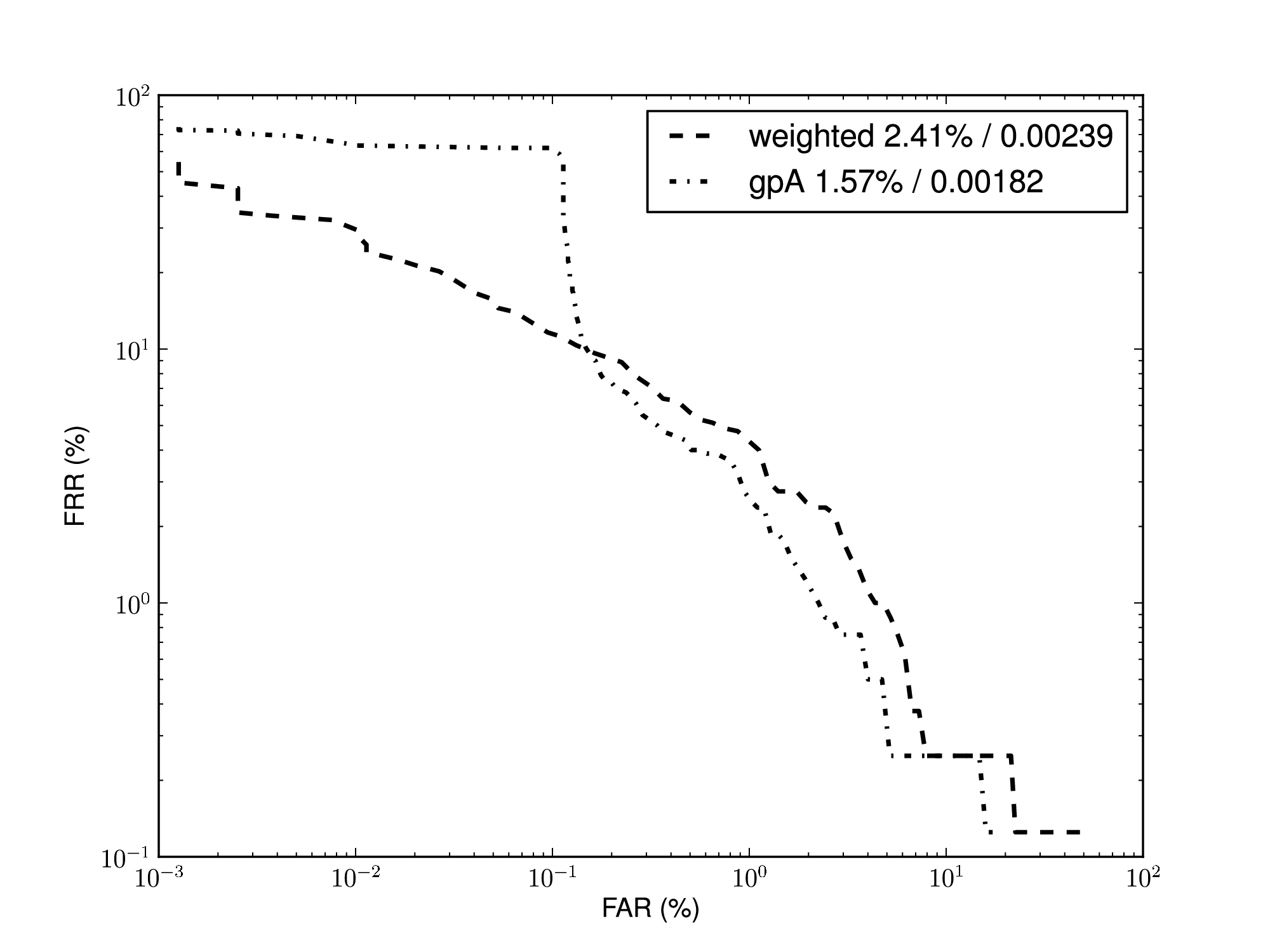}}

  \subfigure[Validation with
  BANCA]{\includegraphics[width=.7\linewidth]{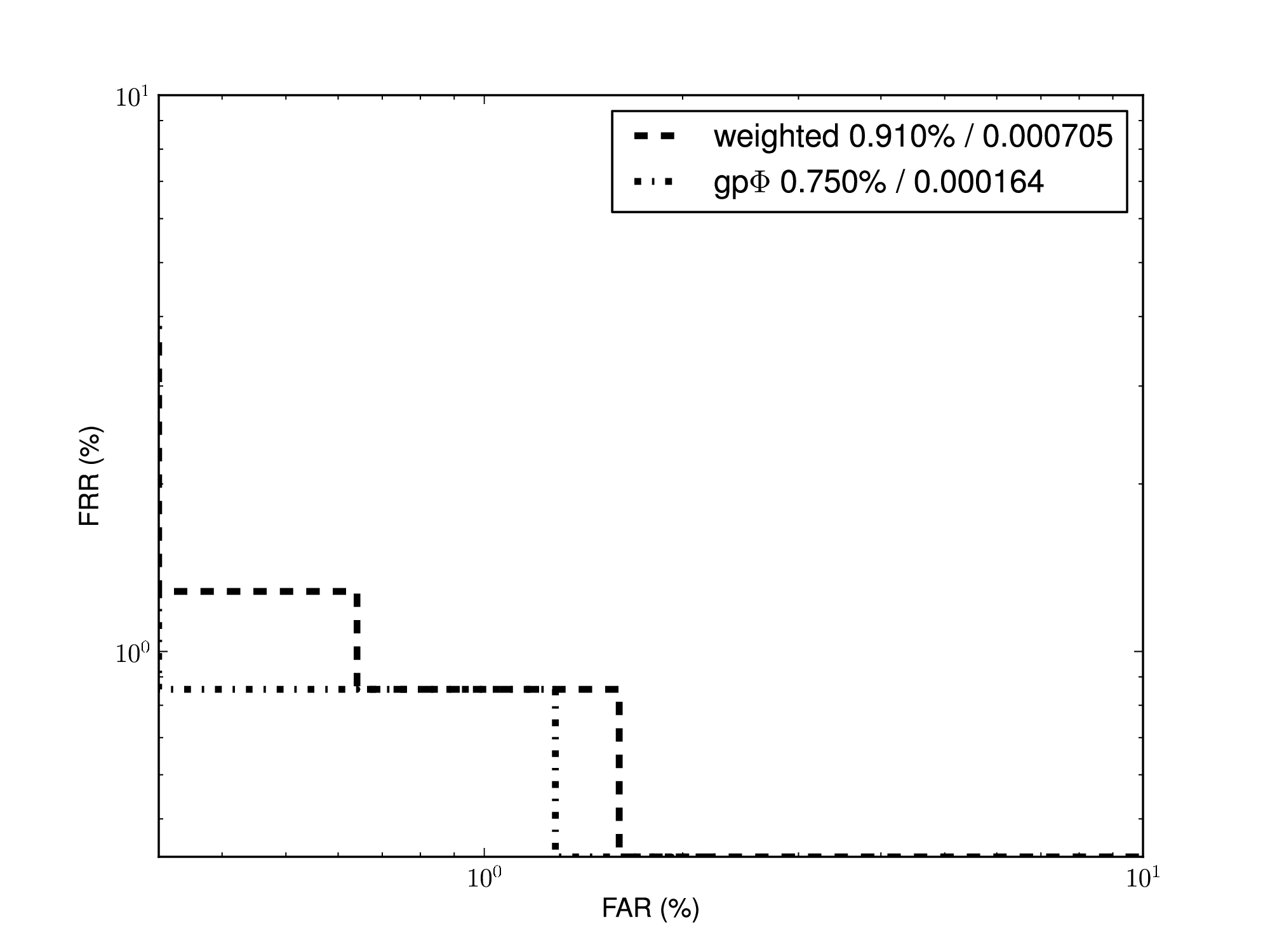}}

  \caption{ROC curves of the fusion systems from the state of the art and with genetic programming.
  The EER of each fusion function is presented in the legend.
  Note the use of a logarithmic scale.}
  \label{fig:roc_systemb}
\end{figure*}

This gain is computed as following: 
\begin{equation}
\label{eq:gain}
gain=100\frac{(EER_{weight} -EER_{gpfunc})}{EER_{weight}}
\end{equation}
where $EER_{weight}$ and $EER_{gpfunc}$ are respectively the EER values of the
weighted fusion and the
generated score fusion function (the same procedure is used for the AUC). 
Better values than the weighted sum are represented in bold.
The EER gives a local performance for one running point (system configured in order
to obtain an FAR equal to the FRR), while the AUC gives a gives a global
performance of the whole system. These two information are really interesting to
use when comparing biometric systems.
Figure~\ref{fig:roc_systemb} presents the ROC curves of the generated programs
against the weighted sum. Performance of the initial biometric
systems are not represented, because we have already seen that they are worst
than the weighted sum (same remark for the other fusion functions). 
Logarithmic scales are used, because error rates are quite small.

We can see from Table~\ref{tab:res2} and Figure~\ref{fig:roc_systemb} that most
of the time, the automatically generated functions with genetic programming give
slightly better results than the weighted sum. These improvements can be local and global
and vary between 16\% and 59\% for the EER and 0.05\% and 76\% for the area
under the curve. When there is no improvement, the results are equal or (in one
case) slightly inferior. Even if there is some difference between training (not
represented in this paper) and
validating sets, we cannot observe overfitting problem.
The BSSR1 dataset presents the largest difference of performance between
training and validation sets, but, the results are still better than the
ones from the state of the art (and the same problem can be observe with the
weighted sum).
By the way, the fitness criterion has never been met, we did not achieve to obtain fusion
functions doing no error. So, the evolution always ended when reaching the $50^{th}$ generation.

\begin{figure*}[!tbh]
 \includegraphics[width=0.9\linewidth]{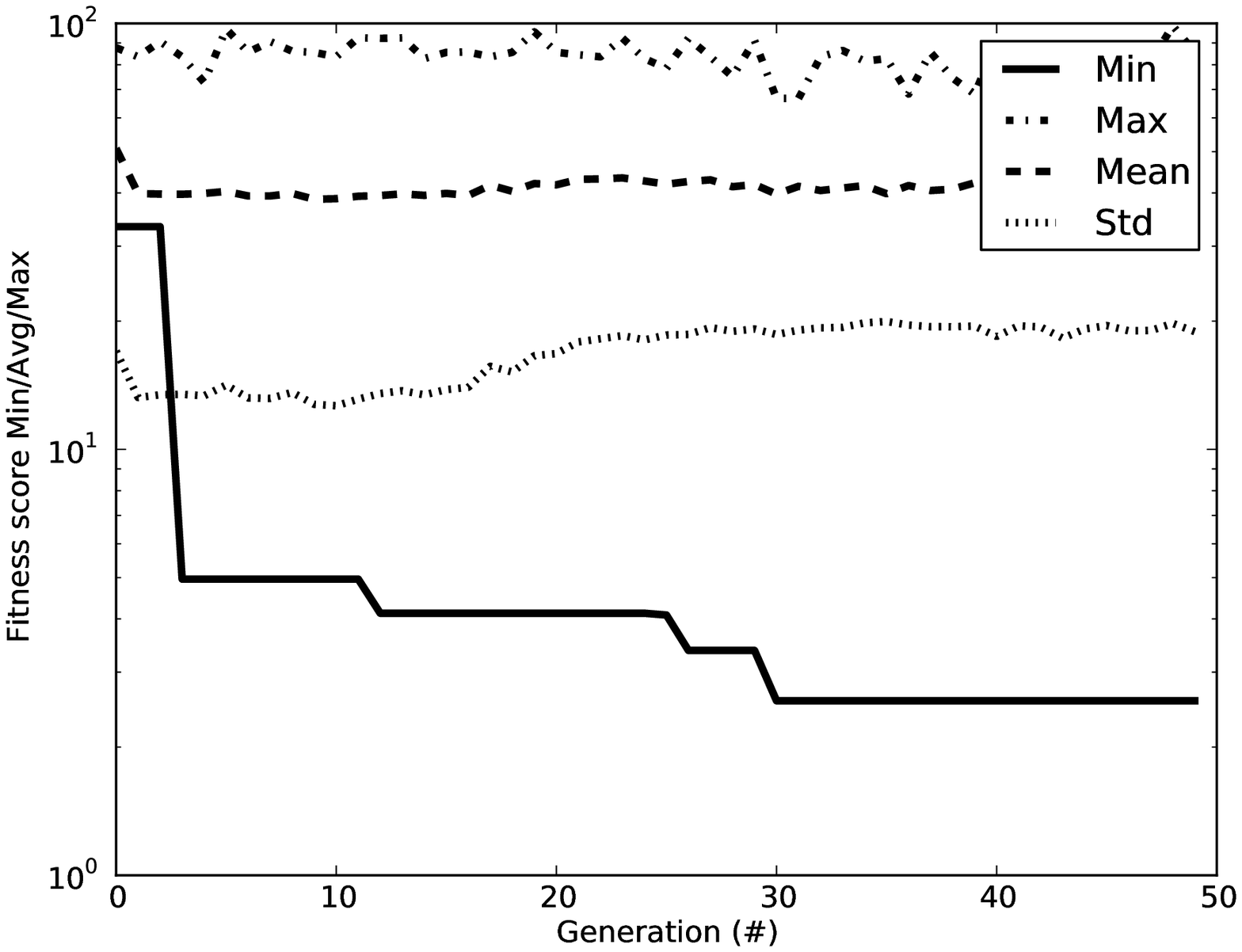}
 \caption{Fitness evolution of one run of the genetic programming
evolution. The max, min, mean and std values of the fitness are represented. We
want to minimize the fitness value, so lower is better.}
 \label{fig:fitness}
\end{figure*}

Figure~\ref{fig:fitness} represents the fitness evolution during all the
generations of one genetic programming run on the BSSR1 database. A logarithmic
scale has been used to give more importance to the low values and track easier
the fitness evolution of the best individual of each generation. We can observe
the same kind of results with the other databases.
The fitness
convergence appears several generations before the end of the computation.
The worst program of each generation is always very
bad which implies that the standard deviation of the fitness is also always quite huge. This
can be explained by the high quantity of mutation probability and the low
quantity of good programs kept for the next generation.
When running the experiment several times, we obtain the same convergence value.
We can say that we reach the maximum performance of the system.

\section{Discussion}

The score fusion functions generated by the proposed approach give a slightly better
performance than the fusion functions used in
the state of the art in multibiometrics.
We can argue that genetic programming is adapted to automatically define score fusion functions returning a
score.
The tradeoff of this performance gain is the need of training patterns which are
not necessary for $sum$, $mul$ or $min$ (but this requirement is already present
for the weighted sum or the use of an SVM). By the way, this is not really a
problem, because we already need training patterns to configure the threshold
of decision (if we do not want to do it empirically) or if we need to normalize
the scores before doing the fusion.

\begin{figure}[!tb]
  \centering
  \includegraphics[width=.9\linewidth]{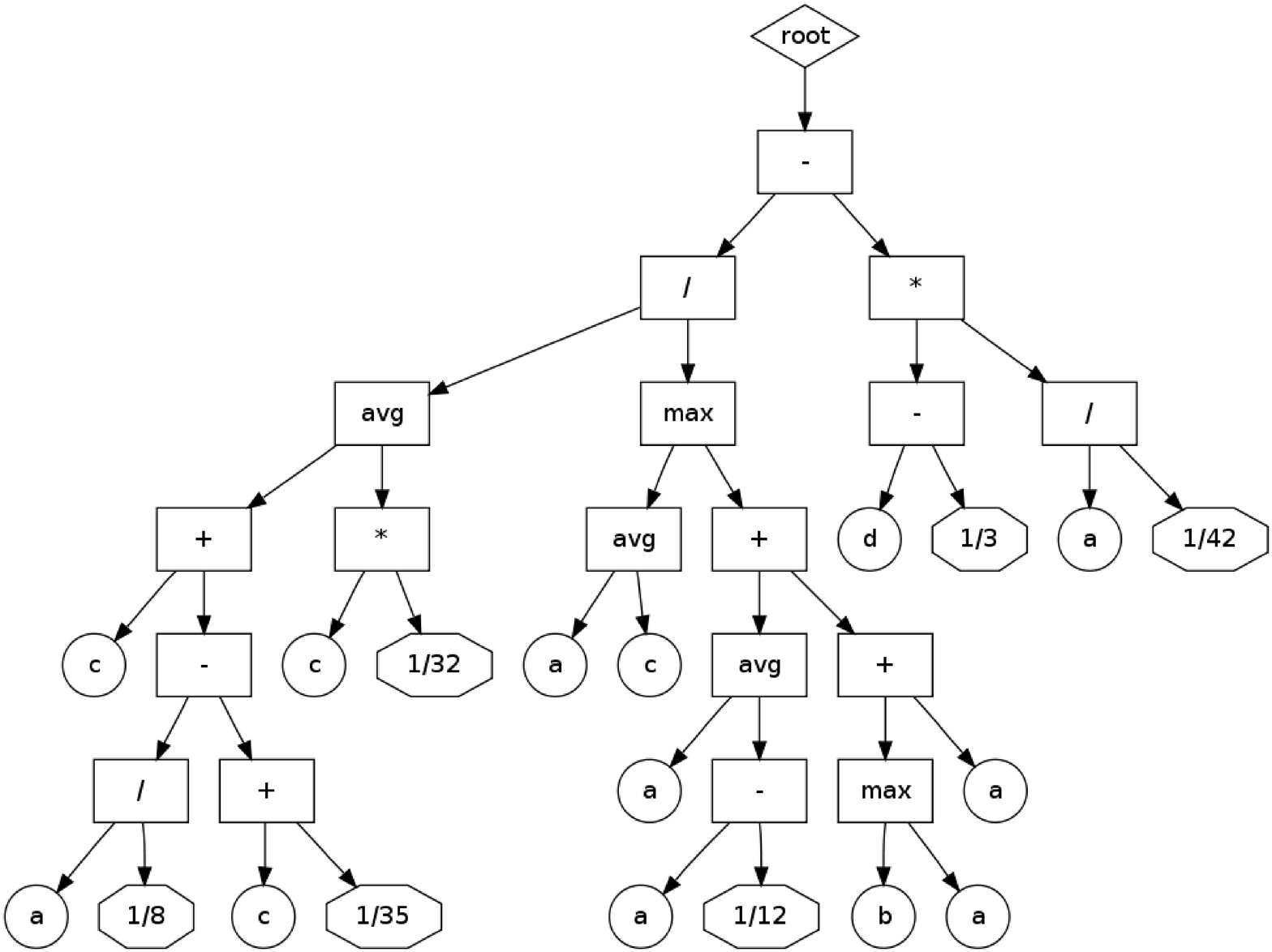}

  \caption{Sample of a "simple" generated program. We can observe the complexity of the
  generated fusion function.}
  \label{fig:arbres}
\end{figure}

Another problem inherent to genetic programming is the complexity of the
generated programs. It is probable that some subtrees could be pruned or
simplified without loosing performance. Another trail would be to add
regularization parameter to the fitness function (for example, the number of
nodes or the depth of the tree). Generated programs would be more
readable by an human and quicker to interpret.
Figure~\ref{fig:arbres} presents a simple generated tree (depending on the
database, they can be more or less complex).
Even if the program is quite short (comparing to
the other generated functions), it includes useless code (e.g., the subtree
$avg(a,a-1/12)$ could be simplified by $a-1/24$).
Some generated trees include preprocessing steps by not using all the
modalities in the terminal set.

Genetic programming generated score fusion functions give performance slightly
equal or better than genetic algorithm configured weighted sum. 
Even if computation time is more important than for genetic algorithm, we
can think that the gain is not really important between the two methods, but, to obtain
these results, genetic programming needed a population ten times smaller and ten times less of generations.

\section{Conclusion}
We propose in this paper a new approach for multibiometrics based on the
automatic generation of 
score fusion functions. We have seen interesting approaches in the state of the art and
decided to improve them by automatically generated score fusion programs by the
help of genetic programming.

Our contribution concerns the designing of multibiometric systems while using a
generic approach based on genetic programming
(and is inspired from the state of the art architectures).
The proposed method returns a multibiometrics score to be 
compared with a defined threshold. 
The proposed multibiometric system has been heavily tested on three
different multibiometric databases. 
We obtained great improvements compared to
classical fusion functions used in the state of the art.
We hope to have opened a new path in the fusion of biometric systems
thanks to genetic programming.

Results could surely be improved by using different parameters in the genetic
programming engine (i.e., more individuals and generations, different range of
constants, different functions, \dots). It could be interesting to test other
performance metrics could be improved by adding quality measures of the capture, and if
genetic programming could produce template fusion programs.


\section{Acknowledgment}
The authors would like to thank: the author of pySTEP~\cite{pySTEP}, the library used during
the experiment, for his helpfull help when encoutering problems with it,
the authors of the various biometric databases used in this experiment,
as well as the French Basse-Normandie region for its financial support.


\begin{thebibliography}{}
\expandafter\ifx\csname url\endcsname\relax
  \def\url#1{\texttt{#1}}\fi
\expandafter\ifx\csname urlprefix\endcsname\relax\def\urlprefix{URL }\fi
\expandafter\ifx\csname href\endcsname\relax
  \def\href#1#2{#2} \def\path#1{#1}\fi

\end{thebibliography}


\begin{thebibliography}{10}
\expandafter\ifx\csname url\endcsname\relax
  \def\url#1{\texttt{#1}}\fi
\expandafter\ifx\csname urlprefix\endcsname\relax\def\urlprefix{URL }\fi
\expandafter\ifx\csname href\endcsname\relax
  \def\href#1#2{#2} \def\path#1{#1}\fi

\bibitem{kumar-human}
A.~Kumar, Y.~Zhou, {Human Identification Using KnuckleCodes}, in: IEEE
  International Conference on Biometrics: Theory, Applications and Systems
  (BTAS 2009), 2009.

\bibitem{Hashiyada2004DoB}
M.~Hashiyada, Developement of biometric dna ink for authentication security,
  Tohoku J. Exp. Med. 204 (2004) 109--117.

\bibitem{Korotkaya2003BPA}
Z.~Korotkaya, Biometric person authentication: Odor, Tech. rep., Department of
  Information Technology, Laboratory of Applied Mathematics, Lappeenranta
  University of Technology (2003).

\bibitem{Riera2008UBS}
A.~Riera, A.~Soria-Frisch, M.~Caparrini, C.~Grau, G.~Ruffini, Unobtrusive
  biometric system based on electroencephalogram analysis, EURASIP Journal on
  Advances in Signal Processing 2008 (2008) 8.

\bibitem{gaines1980}
R.~Gaines, W.~Lisowski, S.~Press, N.~Shapiro, Authentication by keystroke
  timing: some preliminary results, Tech. rep., Rand Corporation (1980).

\bibitem{Fierrez2008HoB}
J.~Fierrez, J.~Ortega-Garcia, On-line signature, Springer US, 2008, pp.
  189--209.

\bibitem{Weiss2007MMB}
A.~Weiss, A.~Ramapanicker, S.~Pranav, S.~Noble, L.~Immohr, Mouse movements
  biometric identification: A feasibility study, in: Proceedings of
  Student/Faculty Research Day, CSIS, Pace University,, 2007.

\bibitem{Petrovska-Delacretaz2007TSV}
D.~Petrovska-Delacretaz, A.~El~Hannani, G.~Chollet, Text-independent speaker
  verification: State of the art and challenges, Lecture Notes In Computer
  Science 4391 (2007) 135.

\bibitem{nandini2008comprehensive}
C.~Nandini, C.~Kumar, Comprehensive framework to gait recognition,
  International Journal of Biometrics 1~(1) (2008) 129--137.

\bibitem{Benli2008DRU}
K.~Benli, R.~Duzagac, M.~Eskil, Driver recognition using gaussian mixture
  models and decision fusion techniques, in: ISICA 2008, 2008.

\bibitem{turk1991face}
M.~Turk, A.~Pentland, Face recognition using eigenfaces, in: Proc. IEEE Conf.
  on Computer Vision and Pattern Recognition, Vol. 591, 1991.

\bibitem{maltoni2009hof}
D.~Maltoni, A.~Jain, S.~Prabhakar, Handbook of fingerprint recognition,
  Springer, 2009.

\bibitem{kumar2006pru}
A.~Kumar, D.~Zhang, Personal recognition using hand shape and texture, IEEE
  Transactions on Image Processing 15~(8) (2006) 2454.

\bibitem{xu5blood}
Z.~Xu, X.~Guo, X.~Hu, X.~Cheng, The blood vessel recognition of ocular fundus,
  in: Proceedings of the 4th International Conference on Machine Learning and
  Cybernetics (ICMLC’05), 2005, pp. 4493--4498.

\bibitem{ross2006handbook}
A.~Ross, K.~Nandakumar, A.~Jain, {Handbook of multibiometrics}, Springer, 2006.

\bibitem{UBESBS08}
M.~Theofanos, B.~Stanton, C.~A. Wolfson, Usability \& Biometrics: Ensuring
  Successful Biometric Systems, National Institute of Standards and Technology
  (NIST), 2008.

\bibitem{isoiec}
ISO, Biometric performance testing and reporting, Tech. rep., ISO/IEC
  1975-1:2006(E) (2006).

\bibitem{Bhatnagar2009}
J.~Bhatnagar, A.~Kumar, On estimating performance indices for biometric
  identification, Pattern Recognition 42 (2009) 1803 -- 1815.

\bibitem{Raghavendra2009PSO}
R.~Raghavendra, B.~Dorizzi, A.~Rao, G.~Hemantha~Kumar, Pso versus adaboost for
  feature selection in multimodal biometrics, in: IEEE 3rd International
  Conference on Biometrics: Theory, Applications and Systems, BTAS 2009, 2009.

\bibitem{rattani2009rma}
A.~Rattani, M.~Tistarelli, Robust multi-modal and multi-unit feature level
  fusion of face and iris biometrics, in: International Conference on
  biometrics (ICB2009), 2009.

\bibitem{ross2004multimodal}
A.~Ross, A.~Jain, Multimodal biometrics: An overview, in: Proceedings of 12th
  European Signal Processing Conference, Citeseer, 2004, pp. 1221--1224.

\bibitem{zuev1999voting}
Y.~Zuev, S.~Ivanov, The voting as a way to increase the decision reliability,
  Journal of the Franklin Institute 336~(2) (1999) 361--378.

\bibitem{buyssens2009}
P.~Buyssens, M.~Revenu, O.~Lepetit, Fusion of ir and visible light modalities
  for face recognition, in: IEEE International Conference on Biometrics:
  Theory, Applications and Systems (BTAS 2009), 2009.

\bibitem{montalvao2006mbf}
J.~Montalvao~Filho, E.~Freire, Multimodal biometric fusion—joint typist
  (keystroke) and speaker verification, in: Telecommunications Symposium, 2006
  International, 2006, pp. 609--614.

\bibitem{Hocquet2007Aba}
S.~Hocquet, Authentification biométrique adaptative application à la
  dynamique de frappe et à la signature manuscrite, Ph.D. thesis, Université
  de Tours (2007).

\bibitem{allano2009}
L.~Allano, La biométrie multimodale : stratégies de fusion de scores et
  mesures de dépendance appliquées aux bases de personnes virtuelles, Ph.D.
  thesis, Institut National des Télécommunications (2009).

\bibitem{shen2009aml}
P.~S. Teh, A.~B.~J. Teoh, C.~Tee, T.~S. Ong, A multiple layer fusion approach
  on keystroke dynamics, Pattern Analysis \& Applications (2009) 14.

\bibitem{nandakumar2008likelihood}
K.~Nandakumar, Y.~Chen, S.~Dass, A.~Jain, {Likelihood ratio-based biometric
  score fusion}, IEEE Transactions on Pattern Analysis and Machine Intelligence
  30~(2) (2008) 342.

\bibitem{Czyz}
J.~Czyz, M.~Sadeghi, J.~Kittler, L.~Vandendorpe, Decision fusion for face
  authentication  7.

\bibitem{garcia2005multimodal}
S.~Garcia-Salicetti, M.~Mellakh, L.~Allano, B.~Dorizzi, Multimodal biometric
  score fusion: the mean rule vs. support vector classifiers, in: Proc.
  EUSIPCO, 2005.

\bibitem{koza1992genetic}
J.~Koza, J.~Rice, {Genetic programming}, Springer, 1992.

\bibitem{ross2009mso}
A.~Ross, N.~Poh, Handbook of Remote Biometrics, Springer, Ch. {Multibiometric
  Systems: Overview, Case Studies, and Open Issues}.

\bibitem{NISTBSSR}
NIST, \href{http://www.itl.nist.gov/iad/894.03/biometricscores/}{Nist biometric
  score set} (2006).
\newline\urlprefix\url{http://www.itl.nist.gov/iad/894.03/biometricscores/}

\bibitem{bssr1url}
N.~I. of~Standards, Technology,
  \href{http://www.itl.nist.gov/iad/894.03/biometricscores/}{Nist biometric
  score set} (2006).
\newline\urlprefix\url{http://www.itl.nist.gov/iad/894.03/biometricscores/}

\bibitem{bailly2003banca}
E.~Bailly-Bailliere, S.~Bengio, F.~Bimbot, M.~Hamouz, J.~Kittler,
  J.~Mari{\'e}thoz, J.~Matas, K.~Messer, V.~Popovici, F.~Por{\'e}e, et~al.,
  {The BANCA database and evaluation protocol}, Lecture Notes in Computer
  Science (2003) 625--638.

\bibitem{bancaurl}
N.~Poh,
  \href{http://info.ee.surrey.ac.uk/Personal/Norman.Poh/web/banca_multi/main.p%
hp?bodyfile=entry_page.html}{Banca score database}.
\newline\urlprefix\url{http://info.ee.surrey.ac.uk/Personal/Norman.Poh/web/ban%
ca_multi/main.php?bodyfile=entry_page.html}

\bibitem{sedgwick2005preliminary}
N.~Sedgwick, C.~Limited, {Preliminary Report on Development and Evaluation of
  Multi-Biometric Fusion using the NIST BSSR1 517-Subject Dataset}, Cambridge
  Algorithmica Linited.

\bibitem{martinez1998ar}
A.~Martinez, R.~Benavente, The ar face database, Tech. rep., CVC Technical
  report (1998).

\bibitem{giot2009benchmark}
R.~Giot, M.~El-Abed, R.~Christophe, Greyc keystroke: a benchmark for keystroke
  dynamics biometric systems, in: IEEE International Conference on Biometrics:
  Theory, Applications and Systems (BTAS 2009), 2009.

\bibitem{lowe2004distinctive}
D.~Lowe, {Distinctive image features from scale-invariant keypoints},
  International journal of computer vision 60~(2) (2004) 91--110.

\bibitem{CI-ROSENBERGER-2008}
C.~Rosenberger, L.~Brun, Similarity-based matching for face authentication, in:
  Proceedings of the International Conference on Pattern Recognition
  (ICPR'2008), Tampa, Florida, USA, 2008.

\bibitem{CI-GIOT-2009-2}
R.~Giot, M.~El-Abed, C.~Rosenberger, Keystroke dynamics with low constraints
  svm based passphrase enrollment, in: IEEE Third International Conference on
  Biometrics : Theory, Applicationsand Systems (BTAS), 2009.

\bibitem{hampel1986robust}
F.~Hampel, E.~Ronchetti, P.~Rousseeuw, W.~Stahel, {Robust statistics: the
  approach based on influence functions}, John Wiley \& Sons New York, 1986.

\bibitem{Jain20052270}
A.~Jain, K.~Nandakumar, A.~Ross,
  \href{http://www.sciencedirect.com/science/article/B6V14-4G0DDW4-1/2/d922960%
ee7ed8928744113dd9494d37a}{Score normalization in multimodal biometric
  systems}, Pattern Recognition 38~(12) (2005) 2270 -- 2285.
\newline\urlprefix\url{http://www.sciencedirect.com/science/article/B6V14-4G0D%
DW4-1/2/d922960ee7ed8928744113dd9494d37a}

\bibitem{mitchell1998introduction}
M.~Mitchell, {An introduction to genetic algorithms}, The MIT press, 1998.

\bibitem{poli2008field}
R.~Poli, W.~Langdon, N.~McPhee, {A field guide to genetic programming}, Lulu
  Enterprises Uk Ltd, 2008, freely available at
  http://www.gp-filed-guide.org.uk.

\bibitem{day2007rts}
P.~Day, A.~K. Nandi, Robust text-independent speaker verification using genetic
  programming, IEEE TRANSACTIONS ON AUDIO, SPEECH, AND LANGUAGE PROCESSING 15
  (2007) 285--295.

\bibitem{pySTEP}
M.~Khoury, \href{http://pystep.sourceforge.net}{Python strongly typed genetic
  programming}.
\newline\urlprefix\url{http://pystep.sourceforge.net}

\bibitem{oliphant2006guide}
T.~Oliphant, {Guide to NumPy}, Spanish Fork, UT, Trelgol Publishing.

\bibitem{CI-GIOT-2010}
R.~Giot, M.~El-Abed, C.~Rosenberger, Fast learning for multibiometrics systems
  using genetic algorithms, in: The International Conference on High
  Performance Computing \& Simulation (HPCS 2010), IEEE Computer Society, Caen,
  France, 2010, p.~8.

\bibitem{CI-GIOT-2010-2}
R.~Giot, B.~Hemery, C.~Rosenberger, Low cost and usable multimodal biometric
  system based on keystroke dynamicsand 2d face recognition, in: IAPR
  International Conference on Pattern Recognition (ICPR), IAPR, Istanbul,
  Turkey, 2010.

\bibitem{chang2001libsvm}
C.~Chang, C.~Lin, {LIBSVM: a library for support vector machines} (2001).

\end{thebibliography}

\end{document}